\documentclass[twoside,leqno,twocolumn]{article}
\usepackage{ltexpprt}
\usepackage{algpseudocode,algorithm,algorithmicx}  
\usepackage{amsmath}
\usepackage{caption} 
\usepackage{xcolor}
\usepackage{makecell}
\usepackage{graphicx}
\PassOptionsToPackage{hyphens}{url}\usepackage{hyperref}
\usepackage{cite}
\usepackage{siunitx}
\usepackage[perpage]{footmisc}
\usepackage{mathtools}
\newcommand\Set[2]{\{\,#1\mid#2\,\}}

\algrenewcommand\algorithmicrequire{\textbf{Inputs:}}  
\algrenewcommand\algorithmicensure{\textbf{Outputs:}}
\renewcommand{\algorithmiccomment}[1]{\bgroup\hfill//~#1\egroup}

\begin{document}

\title{\Large LSCP: Locally Selective Combination in Parallel Outlier Ensembles}

\author{
Yue Zhao\thanks{Department of Computer Science, University of Toronto. \newline
Email: \{yuezhao, znasrullah\}@cs.toronto.edu} \\
\and 
Zain Nasrullah\footnotemark[1] \\
\and
Maciej K. Hryniewicki\thanks{Data Analytics, PricewaterhouseCoopers Canada. \newline
Email: maciej.k.hryniewicki@pwc.com} \\
\and
Zheng Li \thanks{Toronto Campus, Northeastern University. \newline
Email: jk\_zhengli@hotmail.com}
}
\date{}

\maketitle


\fancyfoot[R]{\scriptsize{Copyright \textcopyright\ 2019 by SIAM\\
Unauthorized reproduction of this article is prohibited}}





\begin{abstract} \small\baselineskip=9pt In unsupervised outlier ensembles, the absence of ground truth makes the combination of base outlier detectors a challenging task. Specifically, existing parallel outlier ensembles lack a reliable way of selecting competent base detectors, affecting accuracy and stability, during model combination. In this paper, we propose a framework---called Locally Selective Combination in Parallel Outlier Ensembles (LSCP)---which addresses the issue by defining a local region around a test instance using the consensus of its nearest neighbors in randomly selected feature subspaces. The top-performing base detectors in this local region are selected and combined as the model's final output. Four variants of the LSCP framework are compared with seven widely used parallel frameworks. Experimental results demonstrate that one of these variants, LSCP\_AOM, consistently outperforms baselines on the majority of twenty real-world datasets. \end{abstract} 

\section{Introduction} Outlier detection methods aim to identify anomalous data objects from the general data distribution and are useful for problems such as credit card fraud prevention and network intrusion detection \cite{Chandola2009}. Since the ground truth is often absent in outlier mining \cite{Aggarwal2013}, unsupervised detection methods are commonly used for this task \cite{Breunig2000,Kriegel2009,Chandola2009}. However, unsupervised approaches are susceptible to generating high false positive and false negative rates \cite{Das2016}.
To improve model accuracy and stability in these scenarios, recent research explores ensemble approaches to outlier detection \cite{Aggarwal2013,Aggarwal2017,Rayana2017,Zimek2014}. Ensemble learning combines multiple base estimators to achieve superior detection performance and reliability when compared to an individual estimator \cite{Dietterich2000,Ho1994}. It is important to ensure that the combination process is robust because constituent estimators, if synthesized inappropriately, may be detrimental to the predictive capability of an ensemble\cite{Rayana2014,Rayana2016}. Similar to prior works in classification \cite{Rayana2017}, an outlier ensemble may be characterized as parallel if detectors are generated independently. In contrast, it may be considered sequential if the detector generation, selection or combination is iterative. 

Model combination is important for parallel ensembles to ensure diversity exists among base detectors; however, existing works have not jointly addressed two key limitations in this process. First, most parallel ensembles generically combine all detectors without considering selection. This limits the benefits of model combination since individual base detectors may not be proficient at identifying all outlier instances \cite{Cruz2018}. For example, prior work has demonstrated that the value of good detectors can be neutralized by the inclusion of poor detectors in a generic averaging framework \cite{Aggarwal2017}. Secondly, data locality is rarely emphasized in the context of detector selection and combination leading to potentially sub-optimal outcomes. While it is acknowledged that certain types of outliers are better identified by local data relationships \cite{VanStein2016}, detectors are often evaluated at a global scale, where all training points are considered, instead of the local region related to a test instance. 

To address the aforementioned limitations, we propose a fully unsupervised framework called \textbf{L}ocally \textbf{S}elective \textbf{C}ombination in \textbf{P}arallel Outlier Ensembles (LSCP) to selectively combine base detectors by emphasizing data locality. The idea is motivated by an established supervised ensemble framework known as Dynamic Classifier Selection (DCS) \cite{Ho1994}. DCS selects the best classifier for each test instance by evaluating base classifier competency at a local scale \cite{Cruz2018}. The rationale behind this is that base classifiers will generally not excel at categorizing all unknown test instances and that an individual classifier is more likely to specialize in a specific local region \cite{Cruz2018,Ko2008}. Similarly, LSCP first defines the local region of a test instance by the consensus of the nearest training points in randomly selected feature subspaces, and then identifies the most competent base detector in this local region by measuring similarity relative to a pseudo ground truth (see \cite{Campos2018,Rayana2016} for examples). To further improve algorithm stability and capacity, ensemble variations of LSCP are proposed where promising base detectors are kept for a second-phase combination instead of using the single most competent detector. As a whole, LSCP is intuitive, stable and effective for unsupervised combination of independent outlier detectors.

\pagebreak
Our technical contributions in this paper are:

\begin{enumerate}
\item We propose a novel combination framework which, to the best of our knowledge, is the first published effort to adapt DCS from supervised classification tasks to unsupervised parallel outlier ensembles.
\item Extensive experiments on 20 real-world datasets show that LSCP consistently yields better performance than existing parallel combination methods.
\item As a general framework, LSCP is formulated to be compatible with different types of base detectors; we demonstrate its effectiveness with a homogeneous pool of Local Outlier Factor \cite{Breunig2000} detectors.
\item We employ various analysis methods to improve model interpretability. First, theoretical explanations and complexity are provided. Second, visualization techniques are used to intuitively explain why LSCP works and when to use it. Third, statistical tests are used to compare experimental results. 
\item Effort has been made to streamline the accessibility of LSCP. All source code, experiment results and figures are shared for reproduction\footnote{Repository: \url{https://github.com/yzhao062/LSCP}}. In addition, the framework has been implemented in the Python Outlier Detection (\texttt{PyOD}) toolbox  with a unified API and detailed documentation \cite{Zhao2019pyod}.
\end{enumerate}

\section{Related Works}
\subsection{Dynamic Classifier Selection and Dynamic Ensemble Selection.}
Dynamic Classifier Selection (DCS) is an established combination framework for classification tasks. The technique was first proposed by Ho et al. in 1994 \cite{Ho1994} and then extended, under the name DCS Local Accuracy, by Woods et al. in 1997 \cite{Woods1997} to select the most accurate base classifier in a local region. The motivation behind this approach is that base classifiers often make distinctive errors and offer a degree of complementarity \cite{Britto2014}. Consequently, selectively combining base classifiers can result in a performance improvement over generic ensembles which use the majority vote of all base classifiers. Subsequent theoretical work by Giacinto and Roli validated that, under certain assumptions, the optimal Bayes classifier could be obtained by selecting non-optimal classifiers \cite{Giacinto2000}. DCS was later expanded by Ko et al. to Dynamic Ensemble Selection (DES) which selects multiple base classifiers for a second-phase combination given each test instance \cite{Ko2008}. By minimizing reliance on a single classifier and delegating the classification task to a group competent classifiers, the algorithm has demonstrated that it is more robust than DCS \cite{Ko2008}. Motivated by these approaches, LSCP adapts dynamic selection to unsupervised outlier detection tasks. It should be noted that LSCP is an extension of our prior work on DCSO \cite{Zhao2018dcso}, a recent dynamic detector combination framework. Comparatively, LSCP refines the local region definition process to achieve a more stable combination mechanism. 

\subsection{Data Locality in Outlier Detection.}
\label{ssec:data_locality}
The relationship among data objects is critical in outlier detection and existing algorithms can roughly be categorized as either global or local \cite{Kriegel2009,Rayana2017,Schubert2014}. The former considers all objects during inference while the latter only considers a local selection of objects \cite{Schubert2014}. In both cases, their applicability is dependent on the structure of the data. Global outlier detection algorithms, for example, offer superior performance when outliers are highly distinctive from the data distribution \cite{Rayana2016} but often fail to identify outliers in the local neighborhoods of high-dimensional data \cite{Breunig2000,VanStein2016}. Global models also struggle with data represented by a mixture of distributions, where global characteristics do not necessarily represent the distribution of objects in local regions \cite{VanStein2016}. To address these limitations, numerous works have explored local algorithms such as Local Outlier Factor (LOF) \cite{Breunig2000}, Local Outlier Probabilities (LoOP) \cite{Kriegel2009} and GLOSS \cite{VanStein2016}. However, data locality is rarely considered in the context of detector combination; instead, most combination methods utilize all training data points, e.g., the weight calculation in weighted averaging \cite{Zimek2014}. LSCP explores both global and local data relationships by training base detectors on the entire dataset and emphasizing data locality during detector combination. 

\subsection{Outlier Detector Combination.} 
\label{ssec:static_algo}
Recently, studying outlier ensembles has become a popular research area \cite{Aggarwal2013,Aggarwal2015,Aggarwal2017,Zimek2014} resulting in numerous popular works including: (i) parallel ensembles such as Feature Bagging \cite{Lazarevic2005} and Isolation Forest \cite{Liu2008}; (ii) sequential methods including CARE \cite{Rayana2017}, SELECT \cite{Rayana2016} and BoostSelect \cite{Campos2018} and (iii) hybrid approaches like BORE \cite{Micenkova2015} and XGBOD \cite{Zhao2018}. When the ground truth is unavailable, combining outlier models is challenging.
Existing unsupervised combination algorithms in parallel ensembles are often \textbf{generic} and \textbf{global} (GG); a list of representative GG methods are described below (see \cite{Aggarwal2013,Aggarwal2015,Aggarwal2017,Zimek2014,Rayana2017, Lazarevic2005} for details): 

\begin{enumerate}
	\item Averaging (\textit{GG\_A}): average scores of all detectors.
	\item Maximization (\textit{GG\_M}): maximum score across all detectors.
	\item Weighted Averaging (\textit{GG\_WA}): weight each base detector when averaging.
	\item Threshold Sum (\textit{GG\_TH}): discard all scores below a threshold and sum over the remaining scores. 
	\item Average-of-Maximum (\textit{GG\_AOM}): divide base detectors into subgroups and take the maximum score for each subgroup. The final score is the average of all subgroup scores.
	\item Maximum-of-Average (\textit{GG\_MOA}): divide base detectors into subgroups and take the average score for each subgroup. The final score is the maximum of all subgroup scores.
	\item Feature Bagging (\textit{GG\_FB}): generate a diversified set of base detectors by training on randomly selected feature subsets. The final score is the average of base detector scores.
\end{enumerate}

As discussed in \S \ref{ssec:data_locality}, GG methods ignore the importance of data locality while evaluating and combining detectors, which may be inappropriate given the characteristics of outliers \cite{Breunig2000,VanStein2016}. Moreover, without a selection process, poor detectors may hurt the overall detection performance of an ensemble \cite{Rayana2016,Rayana2017}. All seven GG algorithms are thus included as baselines.

There have been attempts to build selective outlier ensembles sequentially in a boosting style. Rayana and Akoglu introduced \textit{SELECT} \cite{Rayana2016} and \textit{CARE} \cite{Rayana2017} to pick promising detectors and exclude the underperforming ones iteratively, which yielded impressive results on both temporal graphs and multi-dimensional outlier data. Campos et al. further extend this idea by proposing an unsupervised boosting strategy \textit{BoostSelect} for outlier ensemble selection \cite{Campos2018}.
As an alternative to sequential selection models, this paper chooses to focus on parallel detector selection which stresses the importance of data locality. Compared with sequential detector selection methods, our approach can select detectors without iteration which may reduce computational cost. 

\begin{figure*}[ht]
	\includegraphics[width=\linewidth]{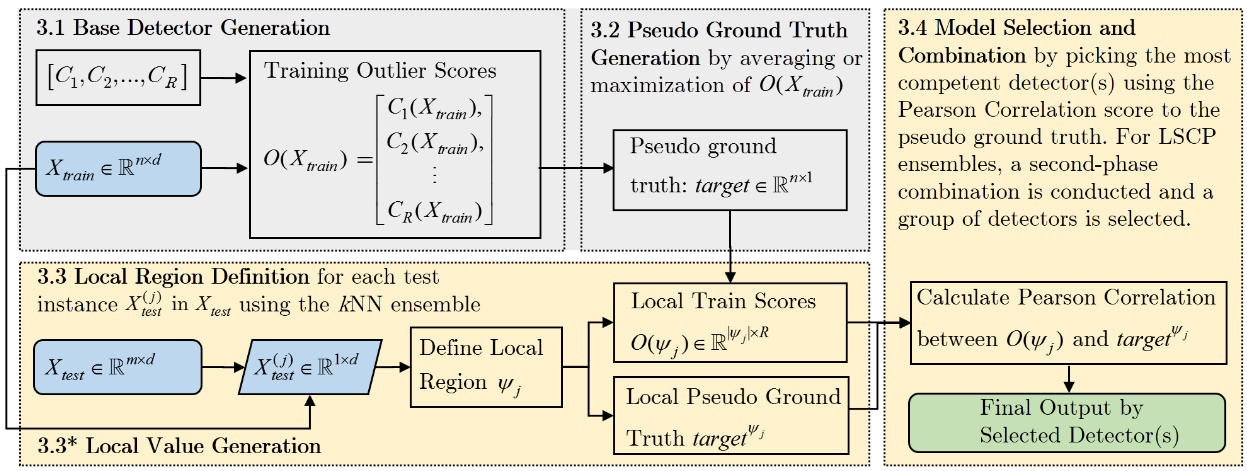}
	\caption{LSCP flow chart. Steps requiring re-computation are highlighted in yellow; cached steps are in gray.}
	\label{fig:flowchart}
\end{figure*} 

\begin{algorithm} [ht]
	\caption{Locally Selective Combination}
	\label{alg:LSCP}
	\begin{algorithmic}[1]
		\Require{the pool of detectors $\mathcal{C}$, training data $X_{train}$, test data $X_{test}$, the local region factor $k$}
		\Ensure{outlier scores of $X_{test}$}
		\State Train all base detectors in $\mathcal{C}$ on $X_{train}$
		\State Get training outlier scores $O(X_{train})$ with Eq.(\ref{e1})
		\State Get pseudo $target\mathrel{\mathop:}=\phi(O(X_{train}))$ with Eq.(\ref{e2})
		\For {each test instance $X_{test}^{(j)}$ in $X_{test}$}
		\State Define local region $\psi_j$ by \textit{k}NN ensemble (\S \ref{ssec:localDef})
		\State Get local pseudo ground truth $target^{\psi_{j}}$ by selecting objects in $\psi_{j}$ from $target$
		\For {each base detector $\mathcal{C}_{r}$ in $\mathcal{C}$}
		\State Get the outlier scores associated with training data in the local region $\mathcal{C}_{r}(\psi_{j})$
		\State Evaluate the local competency of $\mathcal{C}_{r}$ by the Pearson correlation between $target^{\psi_{j}}$ and $\mathcal{C}_{r}(\psi_{j})$
		\EndFor
		\If{\textit{LSCP\_A} or \textit{LSCP\_M}}
		\State \Return $\mathcal{C}^{*}_{r}(X_{test}^{(j)})$ where $\mathcal{C}^{*}_{r}$ has the highest Pearson correlation to $target^{\psi_{j}}$
		\Else \Comment{\textit{LSCP} ensembles} 
		\State Select a group of most similar detectors and add to the empty set $\mathcal{C}^{*}_s$ (\S \ref{ssec:ensemble})
		\If{\textit{LSCP\_AOM}}
        \State \Return $avg(\mathcal{C}^{*}_s(X_{test}^{(j)}))$
		\Else 
        \State \Return $max(\mathcal{C}^{*}_s(X_{test}^{(j)}))$
		\EndIf 
		\EndIf
		\EndFor
	\end{algorithmic}
\end{algorithm}

\section{Algorithm Design}
\label{sec:LSCP}
LSCP starts with a group of diversified detectors to be combined. For each test instance, LSCP first defines its local region and then picks the most competent local detector(s). The selected detector(s) are used to generate the outlier score for the test instance. The workflow of all four proposed LSCP methods is shown in Algorithm \ref{alg:LSCP} and Fig. \ref{fig:flowchart}. 

\subsection{Base Detector Generation.}
An effective ensemble should be constructed with diversified base estimators \cite{Rayana2017,Zimek2014} to promote learning distinct characteristics in the data. With a group of homogeneous base detectors, diversity can be induced by subsampling the training set and feature space, or by varying model hyperparameters \cite{Britto2014,Zimek2014}. In this study, we demonstrate the effectiveness of LSCP by using distinct hyperparameters to construct a pool of models with the same base algorithm. However, in practice, LSCP can also be used as a general framework with heterogeneous base detectors.

Let $X_{train} \in \mathcal{R}^{n\times d}$ denote training data with $n$ points and $d$ features, and $X_{test} \in \mathcal{R}^{m\times d}$ denote a test set with $m$ points. The algorithm first generates a pool of base detectors $\mathcal{C} = \{\mathcal{C}_1,\dots, \mathcal{C}_R\}$ initialized with a range of hyperparameters, e.g., a group of LOF detectors with distinct $MinPts$ \cite{Breunig2000}. All base detectors are first trained on $X_{train}$ and then inference is performed on the same dataset. The results are combined into an outlier score matrix $O(X_{train})$, formalized in Eq.(\ref{e1}), where $\mathcal{C}_r(\cdot)$ denotes the score vector from the $r^{th}$ base detector. Each detector score $\mathcal{C}_{r}(X_{train})$ is normalized using \textit{Z-normalization} as per prior work\cite{Aggarwal2015,Zimek2014}.
\begin{equation} \tag{1}\label{e1}
O(X_{train}) = [\mathcal{C}_{1}(X_{train}),\dots, \mathcal{C}_{r}(X_{train})] \in \mathcal{R}^{n\times R}
\end{equation}

\subsection{Pseudo Ground Truth Generation.} Since LSCP evaluates detector competency without ground truth labels, two methods are used for generating a pseudo ground truth (denoted $target$) with $O(X_{train})$: (i) \textit{LSCP\_A}: averages base detector scores 
and (ii) \textit{LSCP\_M}: maximum score across detectors. This is generalized in Eq.(\ref{e2}) where $\phi$ represents the aggregation (average or max) taken across all base detectors. 
\begin{equation} \tag{2}\label{e2}
target = \phi(O(X_{train})) \in \mathcal{R}^{n\times 1}
\end{equation}
It should be noted that the pseudo ground truth in LSCP is generated using training data and used solely for detector selection.

\subsection{Local Region Definition.} 
\label{ssec:localDef}
The local region $\psi_j$ of a test instance $X_{test}^{(j)}$ is defined as the set of its \textit{k} nearest training objects. Formally, this is denoted as:
\begin{equation} \tag{3}\label{e3}
\psi_j = \Set{x_i}{x_i \in X_{train}, x_i \in kNN_{ens}^{(j)}}
\end{equation}

where $kNN_{ens}$ describes the set of a test instance's nearest neighbours subject to an ensemble criteria. This variation of \textit{k}NN, which is similar to Feature Bagging \cite{Lazarevic2005}, is proposed to alleviate concerns involving the curse of dimensionality on \textit{k}NN \cite{Akoglu2012} while leveraging its better precision compared to clustering algorithms in DCS \cite{Cruz2018}. The process is as follows: (i) $t$ groups of $[\frac{d}{2},d]$ features are randomly selected to construct new feature spaces, (ii) the $k$ nearest training objects to $X_{test}^{(j)}$ in each group are identified using euclidean distance and (iii) training objects that appear more than $\frac{t}{2}$ times are added to $kNN_{ens}^{(j)}$ thus defining the local region. The size of the region is not fixed because it is dependent on the number of training objects that meet the selection criteria.

The local region factor \textit{k} decides the number of nearest neighbors to consider during this process; care is given to avoid selecting extreme values. Smaller values of \textit{k} give more attention to local relationships which can result in instability, while large values of \textit{k} may place too much emphasis on global relationships and have higher computational costs. 
While it is possible to experimentally determine an optimal \textit{k} with cross-validation \cite{Ko2008} when ground truth is available, a similar trivial approach does not exist in an unsupervised setting. For these reasons, we recommend setting $k = 0.1n$, 10\% of the training samples, bounded in the range of $[30,100]$, which yielded good results in practice.

\subsection{Model Selection and Combination.} 

For each test instance, the local pseudo ground truth $target^{\psi_j}$ can be obtained by retrieving values associated with the local region $\psi_j$ from $target$:
\begin{equation} \tag{4}\label{e4}
target^{\psi_j} = \{target_{x_i} | x_i \in \psi_j\} \in \mathcal{R}^{|\psi_j|\times 1}
\end{equation}
where $|\psi_j|$ denotes the cardinality of $\psi_j$. Similarly, the local training outlier scores $O(\psi_j)$ can be retrieved from the pre-calculated training score matrix $O(X_{train})$ as:
\begin{equation} \tag{5}\label{e5}
O(\psi_j) = [\mathcal{C}_{1}(\psi_j),\dots, \mathcal{C}_{r}(\psi_j)] \in \mathcal{R}^{|\psi_{j}| \times R}
\end{equation}

Consequently, although the local region needs to be re-computed for each test instance, the local outlier scores and targets can be efficiently retrieved from pre-calculated values (see Fig. \ref{fig:flowchart}). 

For evaluating base estimator competency in a local region, DCS measures the accuracy of base classifiers as the percentage of correctly classified points \cite{Ko2008}, while LSCP measures the similarity between base detector scores and the pseudo target instead. This distinction is motivated by the lack of direct and reliable ways to access binary labels in unsupervised outlier mining. Although converting pseudo outlier scores to binary labels is feasible, defining an accurate conversion threshold is challenging. Additionally, since imbalanced datasets are common in outlier detection tasks, it is more stable to use similarity measures over absolute accuracy for competency evaluation. Therefore, LSCP measures the local competency of each base detector by the Pearson correlation, which has proven useful in outlier ensemble model combination \cite{Schubert2012evaluation}, between the local pseudo ground truth $target^{\psi_{j}}$ and the local detector score $\mathcal{C}_r(X_{train}^{\psi_j})$. 
The detector $\mathcal{C}^{*}_{r}$ with the highest similarity is regarded as the most competent local detector for $X_{test}^{(j)}$, and its outlier score $\mathcal{C}^{*}_{r}(X_{test}^{(j)})$ can be considered the final score for the corresponding test sample.  


\subsection{Dynamic Outlier Ensemble Selection.}
\label{ssec:ensemble}
Selecting only one detector, even if it is most similar to the pseudo ground truth, can be risky in unsupervised learning. This risk can be mitigated by selecting a group of detectors for a second-phase combination. This idea can be viewed as an adaption of supervised DES \cite{Ko2008} to outlier detection tasks; correspondingly, we introduce ensemble variations of LSCP which employ Maximum of Average (\textit{LSCP\_MOA}) and Average of Maximum (\textit{LSCP\_AOM}) ensembling methods. Specifically, when the pseudo ground truth is generated by $\phi_{average}$, \textit{LSCP\_\textbf{M}OA} selects a subset of competent detectors in the local region of a test instance and then takes the \textbf{m}aximum of their predictions as the outlier score. Inversely, \textit{LSCP\_\textbf{A}OM} computes the \textbf{a}verage of the selected subset when the pseudo target is generated with $\phi_{max}$. Setting the group size of selected detectors equal to 1 is a special case of the ensembles yielding the original LSCP algorithms (\textit{LSCP\_A} and \textit{LSCP\_M}). Larger group sizes may be considered more global in their detector selection while a group size of $R$ results in a fully global algorithm. In response to this, we recommend using a group size selection process which includes some variance. Specifically, a histogram of detector Pearson correlation scores (to the pseudo ground truth) is built with $b$ equal intervals. The detectors belonging to the most frequent interval are kept for the second-phase combination. A large $b$ results in selecting fewer detectors which flexibly controls the strength of the group size in LSCP ensembles. 

The time complexity for training each base detector and generating the pseudo ground truth is dependent on the underlying model and number of training samples. However, since this work proposes a combination framework, our discussion focuses on the overhead introduced during the combination stage. With the appropriate implementation of \textit{LSCP\_A} and \textit{LSCP\_M}, e.g., using a k-d tree, the additional time complexity for defining the local region of each test instance is $O(nd+nlog(n))$: $O(nd)$ for the distance calculation and $O(nlog(n))$ for summation and sorting \cite{Ko2008}. Here, $n$ refers to each test instance and $d$ refers to its dimensionality. While defining the local region requires multiple iterations, the number of iterations is fixed and is thus excluded in the complexity analysis. To combine the $s$ base detectors in \textit{LSCP\_MOA} and \textit{LSCP\_AOM}, an additional $O(s)$ is needed resulting in a total time complexity of $O(nd+nlog(n)+s)$.

\subsection{Theoretical Considerations.}
\label{ssec:theory}
Recently, Aggarwal and Sathe laid the theoretical foundation for outlier ensembles \cite{Aggarwal2015} using the bias-variance tradeoff, a widely used framework for analyzing generalization error in classification problems. 
The reducible generalization error in outlier ensembles may be minimized by either reducing squared bias or variance where a tradeoff between these two channels usually exists.  A high variance detector is sensitive to data variation with high instability; a high bias detector is less sensitive to data variation but may fit complex data poorly. The goal of outlier ensembles is to control both bias and variance to reduce the overall generalization error. Various newly proposed algorithms have been analyzed using this new framework to enhance interpretability \cite{Rayana2016,Rayana2017,Zhao2018}. 

It has been shown that combining diversified base detectors, by averaging them for example, results in variance reduction \cite{Rayana2016,Rayana2017,Aggarwal2015}. However, a combination of all base detectors may also include inaccurate ones leading to higher bias. This explains why generic global averaging does not work well.
Within Aggarwal's bias-variance framework, LSCP possesses a combination of both variance and bias reduction. It induces diversity by initializing various base detectors with distinct hyperparameters and indirectly promotes variance reduction in the way that the pseudo ground truth is generated, e.g., averaging in \textit{LSCP\_A}. Furthermore, LSCP focuses on detector selection by local competency, which helps identify base detectors with conditionally low model bias. \textit{LSCP\_M} is also expected to be more stable than global maximization (\textit{GG\_M}) since the variance is reduced by using the most competent detector's output rather than the global maximum of all base detectors. \textit{LSCP\_MOA} and \textit{LSCP\_AOM} further decrease generalization error through bias and variance reduction, respectively, in their second-phase combination. However, LSCP is still a heuristic framework and can yield unpredictable results on pathological datasets. 

\section{Numerical Experiments}
\subsection{Datasets and Evaluation Metrics.} Table \ref{table:data} summarizes the 20 public outlier detection benchmark datasets used in this study from \texttt{ODDS}\footnote{ODDS Library: \url{http://odds.cs.stonybrook.edu}} and \texttt{DAMI}\footnote{DAMI Datasets: \url{http://www.dbs.ifi.lmu.de/research/outlier-evaluation/DAMI}}. In each experiment, 60\% of the data is used for training and the remaining 40\% is set aside for validation. Performance is evaluated by taking the average score of 30 independent trials using area under the receiver operating characteristic (ROC-AUC) and mean average precision (mAP). Both metrics are widely used in outlier research \cite{Aggarwal2017,Akoglu2012,Micenkova2015,Rayana2016,Zhao2018,Emmott2015} and statistical measures are used to analyze the results \cite{Demsar2006}. Specifically, we use a non-parametric Friedman test followed by a post-hoc Nemenyi test. For these tests, $p<0.05$ is considered to be statistically significant. 

\begin{table}
\centering
	\caption{Real-world datasets used for evaluation} 
	\footnotesize
	\begin{tabular}{l | r  r | r r } 
		\hline 
		\textbf{Dataset} & \textbf{Pts} & \textbf{Dim} & \textbf{Outliers}	& \textbf{\%Outlier}\\
		\hline
		Annthyroid & 7200  & 6   & 534   & 7.41 \\
		Arrhythmia & 452   & 274 & 66   & 14.60 \\
		Breastw    & 683   & 9   & 239  & 34.99 \\
		Cardio     & 1831  & 21  & 176  & 9.61  \\
		Letter     & 1600  & 32  & 100  & 6.25  \\
		MNIST      & 7603  & 100 & 700  & 9.21  \\
		Musk       & 3062  & 166 & 97   & 3.17  \\
		PageBlocks & 5393  & 10  & 510  & 9.46  \\
		Pendigits  & 6870  & 16  & 156  & 2.27  \\
		Pima       & 768   & 8   & 268  & 34.90 \\
		Satellite  & 6435  & 36  & 2036 & 31.64 \\
		Satimage-2 & 5803  & 36  & 71   & 1.22  \\
		Shuttle    & 49097 & 9   & 3511 & 7.15  \\
		SpameSpace & 4207  & 57  & 1679 & 39.91 \\
		Stamps     & 340   & 9   & 31   & 9.12  \\
		Thyroid    & 3772  & 6   & 93   & 2.47  \\
		Vertebral  & 240   &  6  & 30   & 12.50 \\
		Vowels     & 1456  & 12  & 50   & 3.43  \\
		WBC        & 378   & 30  & 21   & 5.56  \\
		Wilt       & 4819  & 5   & 257  & 5.33  \\
		\hline
	\end{tabular}
	\label{table:data} 
\end{table}

\subsection{Experimental Design.}
This study compares the seven GG algorithms introduced in \S \ref{ssec:static_algo} with the four proposed LSCP variations described in Algorithm \ref{alg:LSCP}. All models use a pool of 50 LOF base detectors ensuring consistency during performance evaluation. To induce diversity among base detectors, distinct initialization hyperparameters---specifically the number of neighbors ($MinPts$) used in each LOF detector---are randomly selected in the range of $[5,200]$. It is noted that a narrower range (i.e., $[10,150]$) is taken for certain datasets due to limitations associated with size or computational cost. For \textit{GG\_AOM} and \textit{GG\_MOA}, the base detectors are divided into 5 subgroups and each group contains 10 base detectors selected without replacement. For Feature Bagging (\textit{GG\_FB}), the base detector is randomly chosen from the same pool of LOF detectors and then trained on 50 individual subsamples with $[\frac{d}{2},d]$ randomly selected features. For all LSCP algorithms, the default hyperparameters mentioned in \S \ref{sec:LSCP} are used. 

\begin{table*}[ht]\centering 
	\caption{ROC-AUC scores (average of 30 independent trials, highest score highlighted in bold)} 
	\footnotesize
	\begin{tabular}{l | c c c c | c c c c c c c} 
		\hline 
		\textbf{Dataset} 
		& \makecell{\textbf{LSCP\_} \\ \textbf{A}} 
		& \makecell{\textbf{LSCP\_} \\ \textbf{MOA}} 
		& \makecell{\textbf{LSCP\_} \\ \textbf{M}} 
		& \makecell{\textbf{LSCP\_} \\ \textbf{AOM}} 
		& \makecell{\textbf{GG\_} \\ \textbf{A}} 
		& \makecell{\textbf{GG\_} \\ \textbf{MOA}} 
		& \makecell{\textbf{GG\_} \\ \textbf{M}} 
		& \makecell{\textbf{GG\_} \\ \textbf{AOM}}
		& \makecell{\textbf{GG\_} \\ \textbf{WA}} 
		& \makecell{\textbf{GG\_} \\ \textbf{TH}} 
		& \makecell{\textbf{GG\_} \\ \textbf{FB}} \\ [0.5ex] 
		\hline 
        Annthyroid & 0.7548 & 0.7590 & 0.7849 & 0.7520 & 0.7642 & 0.7660 & 0.7769 & 0.7730 & 0.7632 & 0.7552 & \textbf{0.7854} \\
        Arrhythmia & 0.7746 & 0.7715 & 0.7729 & \textbf{0.7763} & 0.7758 & 0.7749 & 0.7656 & 0.7690 & 0.7758 & 0.7313 & 0.7709 \\
        Breastw    & 0.6553 & 0.7044 & 0.7236 & \textbf{0.7845} & 0.7362 & 0.7140 & 0.6590 & 0.6838 & 0.7453 & 0.6285 & 0.3935 \\
        Cardio     & 0.8691 & 0.8908 & 0.8491 & \textbf{0.9013} & 0.8770 & 0.8865 & 0.8798 & 0.8903 & 0.8782 & 0.8830 & 0.8422 \\
        Letter     & 0.7818 & 0.7954 & 0.8361 & 0.7867 & 0.7925 & 0.8031 & \textbf{0.8434} & 0.8300 & 0.7908 & 0.8001 & 0.7640 \\
        MNIST      & 0.8576 & 0.8623 & 0.7812 & \textbf{0.8633} & 0.8557 & 0.8588 & 0.8349 & 0.8553 & 0.8563 & 0.8272 & 0.8468 \\
        Musk       & 0.9950 & 0.9970 & 0.9931 & \textbf{0.9981} & 0.9937 & 0.9960 & 0.9960 & 0.9970 & 0.9953 & 0.9958 & 0.7344 \\
        PageBlocks & 0.9349 & 0.9343 & 0.8687 & \textbf{0.9488} & 0.9443 & 0.9440 & 0.9240 & 0.9371 & 0.9453 & 0.9418 & 0.9284 \\
        Pendigits  & 0.8238 & 0.8656 & 0.7238 & \textbf{0.8744} & 0.8378 & 0.8509 & 0.8488 & 0.8622 & 0.8425 & 0.8548 & 0.8034 \\
        Pima       & 0.7059 & 0.6991 & 0.6640 & \textbf{0.7061} & 0.7030 & 0.7003 & 0.6730 & 0.6856 & 0.7037 & 0.6349 & 0.6989 \\
        Satellite  & 0.5814 & 0.6106 & 0.6006 & 0.6015 & 0.5881 & 0.5992 & \textbf{0.6258} & 0.6220 & 0.5876 & 0.6101 & 0.5818 \\
        Satimage-2 & 0.9852 & 0.9931 & 0.9878 & \textbf{0.9935} & 0.9872 & 0.9907 & 0.9909 & 0.9925 & 0.9880 & 0.9881 & 0.9181 \\
        Shuttle    & 0.5392 & 0.5551 & 0.5373 & 0.5514 & 0.5439 & 0.5504 & \textbf{0.5612} & 0.5602 & 0.5413 & 0.5561 & 0.3702 \\
        SpamSpace  & 0.3792 & 0.4594 & 0.4305 & \textbf{0.4744} & 0.4487 & 0.4377 & 0.4060 & 0.4128 & 0.4580 & 0.4104 & 0.3312 \\
        Stamps     & 0.8888 & 0.8719 & 0.8525 & \textbf{0.8985} & 0.8946 & 0.8927 & 0.8559 & 0.8763 & 0.8953 & 0.8904 & 0.8715 \\
        Thyroid    & 0.9579 & 0.9624 & 0.9413 & \textbf{0.9700} & 0.9656 & 0.9647 & 0.9385 & 0.9510 & 0.9665 & 0.9644 & 0.8510 \\
        Vertebral  & 0.3324 & 0.3662 & \textbf{0.4306} & 0.3478 & 0.3433 & 0.3467 & 0.3662 & 0.3614 & 0.3442 & 0.3678 & 0.3385 \\
        Vowels     & 0.9276 & 0.9185 & 0.9238 & 0.9199 & 0.9265 & 0.9275 & \textbf{0.9313} & 0.9271 & 0.9261 & 0.9299 & 0.9148 \\
        WBC        & 0.9379 & 0.9344 & 0.9242 & \textbf{0.9451} & 0.9421 & 0.9409 & 0.9321 & 0.9367 & 0.9420 & 0.9314 & 0.9407 \\
        Wilt       & 0.5275 & 0.5517 & \textbf{0.6550} & 0.4286 & 0.5101 & 0.5358 & 0.6384 & 0.6056 & 0.5037 & 0.5586 & 0.5868 \\
		\hline 
	\end{tabular}
	\label{table:roc} 
\end{table*}

\subsection{Algorithm Performances.}

Tables \ref{table:roc} and \ref{table:ap} summarize the ROC-AUC and mAP scores on the 20 datasets. Our experiments demonstrate that LSCP can bring consistent performance improvement over its GG counterparts, which is especially noticeable in the mAP scores. The Friedman test shows there is a statistically significant difference between the 11 algorithms in both ROC-AUC $(\chi^2=43.34, p=\num{4.3160e-6})$ and mAP $(\chi^2=43.49, p=\num{4.0632e-6})$; however, the Nemenyi test fails to identify which pairs of algorithms are significantly different. The latter result is expected in an unsupervised setting due to the difficulty of this task relative to the limited number of datasets \cite{Demsar2006}. In general, LSCP algorithms show great potential: they achieve the highest ROC-AUC scores on 15 datasets and the highest mAP scores on 18 datasets. \textit{GG\_M} yields higher ROC-AUC on four datasets while \textit{GG\_FB} and \textit{GG\_AOM} perform better on \textbf{Annthyroid} in terms of ROC-AUC and mAP respectively. In all other cases, GG algorithms are outperformed by a variant of LSCP. Specifically, \textit{LSCP\_AOM} is the best performing method and ranks highest on 13 datasets in terms of ROC-AUC and 14 datasets in terms of mAP. It should be noted that GG methods with a second-phase combination (\textit{GG\_MOA} and \textit{GG\_AOM}) demonstrate better performance than \textit{GG\_A} and better stability than \textit{GG\_M}. For instance, \textit{GG\_M} has lower mAP than \textit{GG\_AOM} on 15 datasets. These observations agree with the conclusions in Aggarwal's work \cite{Aggarwal2015,Aggarwal2017}.

\begin{table*}[ht]\centering
	\caption{mAP scores (average of 30 independent trials, highest score highlighted in bold)} 
	\footnotesize
	\begin{tabular}{l | c c c c | c c c c c c c} 
		\hline 
		\textbf{Dataset} 
		& \makecell{\textbf{LSCP\_} \\ \textbf{A}} 
		& \makecell{\textbf{LSCP\_} \\ \textbf{MOA}} 
		& \makecell{\textbf{LSCP\_} \\ \textbf{M}} 
		& \makecell{\textbf{LSCP\_} \\ \textbf{AOM}} 
		& \makecell{\textbf{GG\_} \\ \textbf{A}} 
		& \makecell{\textbf{GG\_} \\ \textbf{MOA}} 
		& \makecell{\textbf{GG\_} \\ \textbf{M}} 
		& \makecell{\textbf{GG\_} \\ \textbf{AOM}}
		& \makecell{\textbf{GG\_} \\ \textbf{WA}} 
		& \makecell{\textbf{GG\_} \\ \textbf{TH}} 
		& \makecell{\textbf{GG\_} \\ \textbf{FB}} \\ [0.5ex] 
		\hline 
        Annthyroid & 0.2283 & 0.2375 & 0.2349 & 0.2453 & 0.2301 & 0.2395 & 0.2413 & \textbf{0.2516} & 0.2306 & 0.2277 & 0.1864 \\
        Arrhythmia & 0.3780 & 0.3744 & 0.3790 & \textbf{0.3796} & 0.3766 & 0.3769 & 0.3690 & 0.3722 & 0.3766 & 0.3468 & 0.3707 \\
        Breastw    & 0.4334 & 0.4766 & 0.4728 & \textbf{0.5655} & 0.4995 & 0.4849 & 0.4249 & 0.4577 & 0.5085 & 0.4366 & 0.2854 \\
        Cardio     & 0.3375 & 0.3960 & 0.3197 & \textbf{0.4117} & 0.3516 & 0.3708 & 0.3666 & 0.3864 & 0.3535 & 0.3629 & 0.3643 \\
        Letter     & 0.2302 & 0.2396 & \textbf{0.3346} & 0.2407 & 0.2388 & 0.2473 & 0.3160 & 0.2867 & 0.2372 & 0.2416 & 0.2193 \\
        MNIST      & 0.3933 & 0.3974 & 0.3353 & \textbf{0.3979} & 0.3911 & 0.3941 & 0.3701 & 0.3896 & 0.3918 & 0.3836 & 0.3928 \\
        Musk       & 0.8478 & 0.8773 & 0.8433 & \textbf{0.9240} & 0.8245 & 0.8718 & 0.8479 & 0.8806 & 0.8608 & 0.8629 & 0.5806 \\
        PageBlocks & 0.5805 & 0.5707 & 0.4684 & \textbf{0.6360} & 0.6043 & 0.6016 & 0.5297 & 0.5733 & 0.6077 & 0.6064 & 0.6094 \\
        Pendigits  & 0.0709 & 0.0893 & 0.0625 & \textbf{0.0944} & 0.0777 & 0.0823 & 0.0834 & 0.0895 & 0.0780 & 0.0832 & 0.0834 \\
        Pima       & 0.5092 & 0.5045 & 0.4716 & \textbf{0.5142} & 0.5089 & 0.5054 & 0.4813 & 0.4920 & 0.5095 & 0.4599 & 0.5094 \\
        Satellite  & 0.4077 & 0.4268 & 0.4223 & 0.4196 & 0.4047 & 0.4139 & \textbf{0.4385} & 0.4352 & 0.4047 & 0.4031 & 0.4049 \\
        Satimage-2 & 0.3477 & 0.6248 & 0.3994 & \textbf{0.6249} & 0.3959 & 0.5089 & 0.5344 & 0.5922 & 0.4159 & 0.4114 & 0.4851 \\
        Shuttle    & 0.1228 & 0.1296 & 0.1167 & \textbf{0.1330} & 0.1297 & 0.1316 & 0.1239 & 0.1294 & 0.1293 & 0.1316 & 0.0549 \\
        SpamSpace  & 0.3326 & 0.3615 & 0.3592 & \textbf{0.3665} & 0.3572 & 0.3521 & 0.3379 & 0.3413 & 0.3612 & 0.3601 & 0.3079 \\
        Stamps     & 0.3596 & 0.3310 & 0.3193 & \textbf{0.3779} & 0.3694 & 0.3660 & 0.3144 & 0.3387 & 0.3706 & 0.3638 & 0.3535 \\
        Thyroid    & 0.3544 & 0.3955 & 0.2638 & \textbf{0.4651} & 0.4045 & 0.4123 & 0.2850 & 0.3488 & 0.4130 & 0.4071 & 0.1186 \\
        Vertebral  & 0.0948 & 0.1020 & \textbf{0.1230} & 0.0988 & 0.0971 & 0.0975 & 0.1029 & 0.1000 & 0.0972 & 0.1067 & 0.0965 \\
        Vowels     & \textbf{0.3913} & 0.3678 & 0.3482 & 0.3539 & 0.3783 & 0.3790 & 0.3760 & 0.3732 & 0.3784 & 0.3783 & 0.3340 \\
        WBC        & 0.6033 & 0.5983 & 0.5472 & \textbf{0.6131} & 0.6097 & 0.6069 & 0.5579 & 0.5925 & 0.6105 & 0.6045 & 0.5933 \\
        Wilt       & 0.0518 & 0.0557 & \textbf{0.0770} & 0.0423 & 0.0493 & 0.0523 & 0.0715 & 0.0633 & 0.0486 & 0.0537 & 0.0591 \\
		\hline 
	\end{tabular}
	\label{table:ap} 
\end{table*}

\begin{figure*}
	\includegraphics[width=\linewidth]{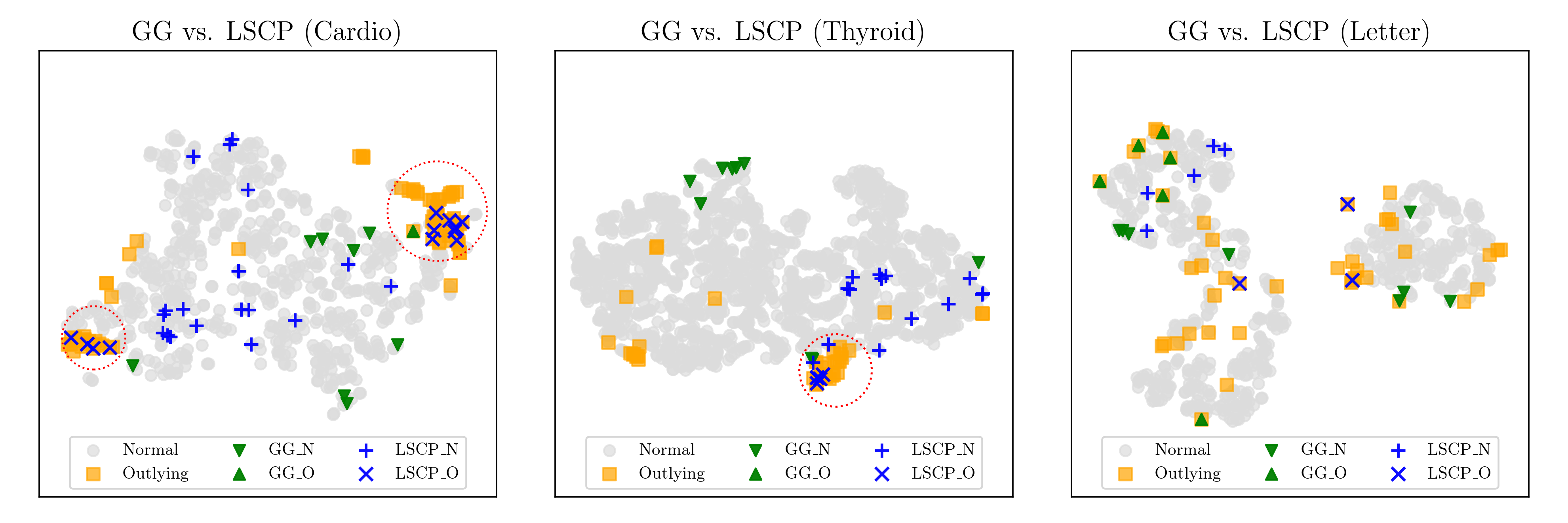}
	\caption{t-SNE visualizations on \textbf{Cardio} (left), \textbf{Thyroid} (middle) and \textbf{Letter} (right), where normal (N) and outlying (O) points are denoted as grey dots and orange squares, respectively. Points that can only be correctly classified by a particular framework are shown in green and blue for GG and LSCP respectively.}
	\label{fig:tsne}
\end{figure*}

\textit{LSCP\_A} and \textit{LSCP\_M} do not demonstrate strong performance relative to their GG counterparts. Given that both pseudo ground truth generation methods are heuristic, they may result in poor local competency evaluation. For example, as discussed in \S \ref{ssec:theory}, \textit{LSCP\_A} theoretically benefits from both the variance and bias reduction by averaging and focusing on locality. In practice though, by only selecting the single most competent detector, it's possible that this approach yields weaker variance reduction compared to \textit{GG\_A} which uses all detector scores. As consequence, the variance reduction may not be able to sufficiently offset the bias inherent to the the pseudo ground truth generation process leading to diminished performance. Comparatively, when the ground truth is generated by taking the maximum among multiple detectors, \textit{LSCP\_M} exhibits unstable behaviour similar to \textit{GG\_M}. \textit{LSCP\_M} only outperforms \textit{LSCP\_AOM} on \textbf{Letter}, \textbf{Vertebral} and \textbf{Wilt} in terms of both ROC-AUC and mAP. As discussed in \cite{Aggarwal2015,Aggarwal2017}, selecting maximum scores across detectors yields high model variance which explains these results; a second-phase combination may mitigate this risk.

A Friedman test confirms there is a significant difference among the four LSCP algorithms in both ROC-AUC $(\chi^2=13.38, p=0.0039)$ and mAP $(\chi^2=21.18, p=\num{9.6592e-5})$. Correspondingly, the LSCP ensemble variation \textit{LSCP\_AOM} shows promise. Building on \textit{GG\_AOM}'s success as one of the most effective combination methods \cite{Aggarwal2017}, \textit{LSCP\_AOM} averages outlier scores within the selected group of detectors which could be viewed as an additional reduction of model variance over \textit{LSCP\_M}. 
Moreover, \textit{LSCP\_AOM}'s concentration on the local competency evaluation may have improved model bias in addition to the variance reduction by the second-phase averaging, leading to better accuracy.
The results show that \textit{LSCP\_AOM} outperforms all models on 13 datasets in terms of ROC-AUC and 14 datasets in terms of mAP. The latter improvement over GG methods is especially considerable on \textbf{Breastw}, \textbf{Cardio}, \textbf{Satimage-2} and \textbf{Thyroid}.

The benefit of taking the second-phase combination is less effective for \textit{LSCP\_MOA}, which does not outperform \textit{LSCP\_A} or \textit{GG\_MOA}. As discussed in \cite{Aggarwal2015,Aggarwal2017}, it is less effective to do a second-phase combination after averaging because information has already been already lost to blunting. The experiment results confirm that, in an LSCP scheme, the benefit from a second-phase maximization cannot offset the information loss due to the initial averaging. Overall, only \textit{LSCP\_AOM} is recommended for detector combination due to its combined bias and variance reduction capability.

\subsection{Visualization Analysis.}
Figure \ref{fig:tsne} visually compares the performance of the best performing GG and LSCP methods on \textbf{Cardio}, \textbf{Thyroid} and \textbf{Letter} using t-Distributed Stochastic Neighbor Embedding (t-SNE) \cite{VanDerMaaten2008}. The green and blue markers highlight objects that can only be correctly classified by GG and LSCP methods respectively to emphasize the mutual exclusivity of the two approaches. The visualizations of \textbf{Cardio} (left) and \textbf{Thyroid} (middle) illustrate that LSCP methods have an edge over GG methods in detecting local outliers when they cluster together (highlighted by red dotted circles in Fig. \ref{fig:tsne}). Additionally, LSCP methods can contribute to classifying both outlying and normal points when locality is present in the data. However, the outlying data distribution in \textbf{Letter} (right) is more dispersed---outliers do not form local clusters but rather mix with normal points. This causes LSCP to perform worse than \textit{GG\_M} in terms of ROC-AUC despite showing an improvement in terms of mAP. Based on these visualizations, one could assume that LSCP is useful when outlying and normal objects are well separated, but less effective when they are interleaved.
Additionally, this suggests that the size of LSCP's local region should be informed by the estimated proportion of outliers in the dataset. For instance, outliers account for only 3.43\% and 6.25\% of \textbf{Vowels} and \textbf{Letter} respectively, which may not be sufficient to form outlier clusters when the local region size is set to 10\% of the training data. A smaller local region size is more appropriate when a small number of outliers is assumed.

\subsection{Limitations and Future Directions.}
Firstly, the local region definition in LSCP depends on finding nearest neighbors by euclidean distance. However, this approach is not ideal due to: (i) high time complexity and (ii) degraded performance when many irrelevant features are presented in high dimensional space \cite{Akoglu2012}. It may be improved by using prototype selection \cite{Cruz2018} or by defining the local region using advanced clustering methods \cite{Cruz2018}. Secondly, only simple pseudo ground truth generation methods are explored (averaging or maximization) in this study; more accurate methods should be considered, such as actively pruning base detectors \cite{Rayana2016} and greedy model combination \cite{Schubert2012evaluation}. Lastly, DCS has proven to work with heterogeneous base classifiers in classification problems \cite{Ko2008,Cruz2018}, which is pending for verification in LSCP. A more significant performance improvement is expected for LSCP framework when the base detectors are more diversified.

\section{Conclusions}
In this work, we propose four variants of a novel unsupervised outlier detection framework called Locally Selective Combination in Parallel Outlier Ensembles (LSCP). Unlike traditional combination approaches, LSCP identifies the top-performing base detectors for each test instance relative to its local region. To validate its effectiveness, the proposed framework is assessed on 20 real-world datasets and is found to be superior to baseline algorithms. The ensemble approach \textit{LSCP\_AOM} demonstrates the best performance achieving the highest detection score on 13/20 datasets with respect to ROC-AUC and 14/20 datasets with respect to mAP. Theoretical considerations under the bias-variance framework and visualizations are also provided for LSCP to provide a holistic overview of the framework. Since LSCP demonstrates the promise of data locality, future work can extend this exploration by investigating the use of heterogeneous base detectors and more reliable pseudo ground truth generation methods.

\bibliography{References}

\begin{thebibliography}{10}

\bibitem{Aggarwal2013}
{\sc C.~C. Aggarwal}, {\em Outlier ensembles: position paper}, ACM SIGKDD
  Explorations, 14 (2013), pp.~49--58.

\bibitem{Aggarwal2015}
{\sc C.~C. Aggarwal and S.~Sathe}, {\em {Theoretical Foundations and Algorithms
  for Outlier Ensembles}}, ACM SIGKDD Explorations, 17 (2015), pp.~24--47.

\bibitem{Aggarwal2017}
\leavevmode\vrule height 2pt depth -1.6pt width 23pt, {\em {Outlier ensembles:
  An introduction}}, Springer, 1st~ed., 2017.

\bibitem{Akoglu2012}
{\sc L.~Akoglu, H.~Tong, J.~Vreeken, and C.~Faloutsos}, {\em {Fast and Reliable
  Anomaly Detection in Categorical Data}}, in CIKM, 2012.

\bibitem{Breunig2000}
{\sc M.~M. Breunig, H.-P. Kriegel, R.~T. Ng, and J.~{\" {o}}.~r. Sander}, {\em
  {LOF: Identifying Density-Based Local Outliers}}, ACM SIGMOD,  (2000),
  pp.~1--12.

\bibitem{Britto2014}
{\sc A.~S. Britto, R.~Sabourin, and L.~E. Oliveira}, {\em {Dynamic selection of
  classifiers - A comprehensive review}}, Pattern Recognition, 47 (2014),
  pp.~3665--3680.

\bibitem{Campos2018}
{\sc G.~O. Campos, A.~Zimek, and W.~Meira}, {\em {An Unsupervised Boosting
  Strategy for Outlier Detection Ensembles}}, PAKDD,  (2018), pp.~564--576.

\bibitem{Chandola2009}
{\sc V.~Chandola, A.~Banerjee, and V.~Kumar}, {\em Anomaly detection: A
  survey}, CSUR, 41 (2009), p.~15.

\bibitem{Cruz2018}
{\sc R.~M. Cruz, R.~Sabourin, and G.~D. Cavalcanti}, {\em {Dynamic classifier
  selection: Recent advances and perspectives}}, Information Fusion, 41 (2018),
  pp.~195--216.

\bibitem{Das2016}
{\sc S.~Das, W.-K. Wong, T.~Dietterich, A.~Fern, and A.~Emmott}, {\em
  {Incorporating Expert Feedback into Active Anomaly Discovery}}, ICDM,
  (2016), pp.~853--858.

\bibitem{Demsar2006}
{\sc J.~Dem~{\v {s}} ar}, {\em {Statistical Comparisons of Classifiers over
  Multiple Data Sets}}, JMLR, 7 (2006), pp.~1--30.

\bibitem{Dietterich2000}
{\sc T.~G. Dietterich}, {\em {Ensemble Methods in Machine Learning}}, MCS, 1857
  (2000), pp.~1--15.

\bibitem{Emmott2015}
{\sc A.~Emmott, S.~Das, T.~Dietterich, A.~Fern, and W.-k. Wong}, {\em {A
  Meta-Analysis of the Anomaly Detection Problem}}, arXiv preprint,  (2015).

\bibitem{Giacinto2000}
{\sc G.~Giacinto and F.~Roli}, {\em {A theoretical framework for dynamic
  classifier selection}}, ICPR, 2 (2000), pp.~0--3.

\bibitem{Ho1994}
{\sc T.~K. Ho, J.~J. Hull, and S.~N. Srihari}, {\em {Decision Combination in
  Multiple Classifier Systems}}, TPAMI, 16 (1994), pp.~66--75.

\bibitem{Ko2008}
{\sc A.~H. Ko, R.~Sabourin, and A.~S. Britto}, {\em {From dynamic classifier
  selection to dynamic ensemble selection}}, Pattern Recognition, 41 (2008),
  pp.~1735--1748.

\bibitem{Kriegel2009}
{\sc H.-P. Kriegel, P.~Kr~{\" {o}} ger, E.~Schubert, and A.~Zimek}, {\em {LoOP:
  local outlier probabilities}}, CIKM,  (2009), pp.~1649--1652.

\bibitem{Lazarevic2005}
{\sc A.~Lazarevic and V.~Kumar}, {\em {Feature bagging for outlier detection}},
  ACM SIGKDD,  (2005), p.~157.

\bibitem{Liu2008}
{\sc F.~T. Liu, K.~M. Ting, and Z.~H. Zhou}, {\em {Isolation forest}}, ICDM,
  (2008), pp.~413--422.

\bibitem{VanDerMaaten2008}
{\sc L.~v.~d. Maaten and G.~Hinton}, {\em Visualizing data using t-sne}, JMLR,
  9 (2008), pp.~2579--2605.

\bibitem{Micenkova2015}
{\sc B.~Micenkov~{\' {a}}, B.~McWilliams, and I.~Assent}, {\em {Learning
  Representations for Outlier Detection on a Budget}}, arXiv preprint,  (2015).

\bibitem{Rayana2014}
{\sc S.~Rayana and L.~Akoglu}, {\em {An Ensemble Approach for Event Detection
  and Characterization in Dynamic Graphs}}, in ACM SIGKDD ODD Workshop, 2014.

\bibitem{Rayana2016}
\leavevmode\vrule height 2pt depth -1.6pt width 23pt, {\em {Less is More:
  Building Selective Anomaly Ensembles}}, TKDD, 10 (2016), pp.~1--33.

\bibitem{Rayana2017}
{\sc S.~Rayana, W.~Zhong, and L.~Akoglu}, {\em {Sequential ensemble learning
  for outlier detection: A bias-variance perspective}}, ICDM,  (2017),
  pp.~1167--1172.

\bibitem{Schubert2012evaluation}
{\sc E.~Schubert, R.~Wojdanowski, A.~Zimek, and H.-P. Kriegel}, {\em On
  evaluation of outlier rankings and outlier scores}, in SDM, SIAM, 2012,
  pp.~1047--1058.

\bibitem{Schubert2014}
{\sc E.~Schubert, A.~Zimek, and H.~P. Kriegel}, {\em {Local outlier detection
  reconsidered: A generalized view on locality with applications to spatial,
  video, and network outlier detection}}, DMKD, 28 (2014), pp.~190--237.

\bibitem{VanStein2016}
{\sc B.~van Stein, M.~van Leeuwen, and T.~B~{\"a} ck}, {\em Local
  subspace-based outlier detection using global neighbourhoods}, IEEE
  International Conference on Big Data,  (2016), pp.~1136--1142.

\bibitem{Woods1997}
{\sc K.~Woods, W.~Kegelmeyer, and K.~Bowyer}, {\em {Combination of multiple
  classifiers using local accuracy estimates}}, TPAMI, 19 (1997), pp.~405--410.

\bibitem{Zhao2018dcso}
{\sc Y.~Zhao and M.~K. Hryniewicki}, {\em {DCSO:} dynamic combination of
  detector scores for outlier ensembles}, in ACM SIGKDD ODD Workshop, London,
  UK, 2018.

\bibitem{Zhao2018}
{\sc Y.~Zhao and M.~K. Hryniewicki}, {\em {XGBOD: Improving Supervised Outlier
  Detection with Unsupervised Representation Learning}}, IJCNN,  (2018).

\bibitem{Zhao2019pyod}
{\sc Y.~Zhao, Z.~Nasrullah, and Z.~Li}, {\em {PyOD}: A python toolbox for
  scalable outlier detection}, arXiv preprint arXiv:1901.01588,  (2019).

\bibitem{Zimek2014}
{\sc A.~Zimek, R.~J. G.~B. Campello, and J.~{\" {o}}.~r. Sander}, {\em
  {Ensembles for unsupervised outlier detection: Challenges and research
  questions}}, ACM SIGKDD Explorations, 15 (2014), pp.~11--22.

\end{thebibliography}
\bibliographystyle{siam}
\end{document}